\documentclass[11pt]{article}

\usepackage[final]{acl}

\usepackage{times}
\usepackage{latexsym}
\usepackage[T1]{fontenc}
\usepackage[utf8]{inputenc}
\usepackage{microtype}
\usepackage{inconsolata}
\usepackage{graphicx}

\usepackage{booktabs}
\usepackage{multirow}
\usepackage[table]{xcolor}
\usepackage{enumitem}
\usepackage{algorithm}
\usepackage{amsmath,amssymb}
\usepackage{algpseudocode}
\usepackage[most]{tcolorbox}
\usepackage{rotating}

\newcommand{\boldpara}[1]{%
  \par\vspace{1ex}\noindent\textbf{#1}\space
}

\definecolor{bestcolor}{HTML}{B5D9D2}
\newcommand{\best}[1]{\cellcolor{bestcolor}\textbf{#1}}

\setlist[itemize]{leftmargin=*}
\setitemize[1]{itemsep=2pt,partopsep=0pt,parsep=\parskip,topsep=2pt}

\title{Don't Box Me In: Dynamic Cultural Adaptation and Cognitive Tracking for Social Understanding}

\author{
\textbf{Chongyuan Dai}$^{1}$,
\textbf{Yaling Shen}$^{2}$,
\textbf{Shengeng Tang}$^{1}$,
\textbf{Hui Ma}$^{1}$,
\textbf{Jinpeng Hu}$^{1\dagger}$\\
    $^1$Hefei University of Technology,
    $^2$Monash University\\
\texttt{2023217261@mail.hfut.edu.cn, jinpenghu@hfut.edu.cn}
}

\begin{document}
\maketitle

\begingroup
    \renewcommand\thefootnote{}
    \footnotetext{$^\dagger$Corresponding author}
\endgroup

\begin{abstract}
Social interaction increasingly takes place in multicultural settings, where individuals may draw on multiple cultural influences and adapt their communicative behavior across contexts.
Despite recent advances in equipping Large Language Models (LLMs) with social understanding capabilities, existing approaches often model culture as a static demographic attribute, limiting their ability to accommodate hybrid and dynamically expressed communicative preferences.
Therefore, in this paper, we propose \textbf{DyCAC}, a training-free framework that achieves fluid social alignment by incorporating \underline{Dy}namic \underline{C}ultural \underline{A}daptation with continuous \underline{C}ognitive tracking. 
Rather than inferring a fixed cultural identity, DyCAC models culturally relevant communicative preferences as a time-varying mixture of population-level cultural reference profiles.
This reference-based representation is further calibrated using dialogue-style signals observed in the ongoing interaction, enabling the model to capture both composite cultural influences and turn-level shifts in communicative behavior.
In parallel, a memory module driven by Theory of Mind (ToM) continuously tracks the cognitive states of the interlocutor.
Extensive experiments on interactive social and cultural benchmarks demonstrate the superiority of our approach.
The proposed framework outperforms existing baselines, exhibiting enhanced social intelligence and broad adaptability across varied multicultural contexts.\footnote{\url{https://github.com/MindIntLab-HFUT/DyCAC}}

\end{abstract}

\begin{figure*}[t]
\centering
\includegraphics[width=0.95\textwidth]{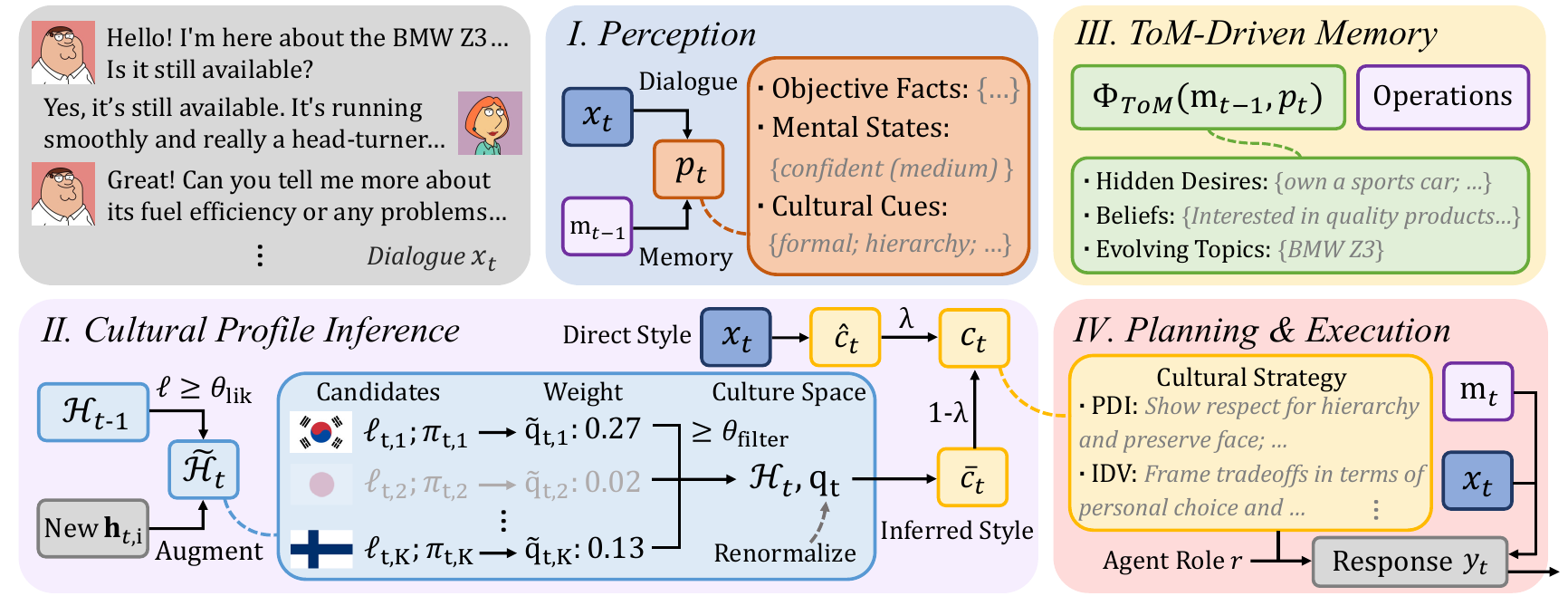} 
\caption{
An overview of the DyCAC framework.
$x_t$ and $y_t$ are the dialogue input and generated response, respectively.
$p_t$ is the perceptual state, and $m_t$ is the memory updated by reasoner $\Phi_{ToM}$.
For cultural inference, variables $\mathcal{H}_t, \pi_{t,i}, \ell_{t,i}, q_{t,i}$ represent the culture space, cultural prior, compatibility score, and normalized weight, respectively.
The final cultural profile $\mathbf{c}_t$ interpolates between the inferred profile $\mathbf{\bar{c}}_t$ and direct style $\mathbf{\hat{c}}_t$ via weight $\lambda$.
}
\label{fig:fig1}
\vspace{-1em}
\end{figure*}

\section{Introduction}
Social interaction fundamentally shapes how humans navigate shared environments, coordinate goal-directed actions, and construct mutual understanding \cite{doi:10.1152/physrev.00041.2007}.
A cornerstone of this capability is social understanding, which is the capacity to maintain coherent mental models of interlocutors and generate responses that align with interpersonal, emotional, and cultural expectations.
Therefore, equipping Large Language Models (LLMs) with robust social abilities has emerged as a pivotal frontier, advancing research across social dialogue \cite{zhou2024sotopia, kong-etal-2025-sdpo, yu2025sotopiarl, zhang-etal-2025-sotopia, bhattacharyya2026social}, emotion understanding \cite{10.1145/3746027.3755726, LIAO2026113366, zheng2026surfacingunsaidcuebenchaffective, song2026bridging}, psychological understanding \cite{10772313, dai-etal-2026-psyche}, and culturally grounded reasoning \cite{alkhamissi-etal-2024-investigating, NEURIPS2024_culturepark, ki-etal-2025-multiple}.
To this end, numerous studies have introduced specific mechanisms, including Theory of Mind (ToM) \cite{kim2025hypothesisdriven, NEURIPS2025_AutoToM} to track cognitive states, and social-oriented training paradigms \cite{wang-etal-2024-sotopiapi, liu-etal-2025-epo} to optimize the interaction process.

Despite this progress, many existing approaches still underemphasize a crucial factor in social interaction: cultural context.
Culture shapes how people interpret politeness, directness, hierarchy, uncertainty, cooperation, and emotional expression \cite{hofstede2011dimensionalizing}.
Therefore, models ignoring cultural context may produce responses that are linguistically fluent but socially misaligned.
Recognizing this gap, recent research has focused on developing culturally aware language models \cite{NEURIPS2024_culturellm, feng-etal-2025-culfit, guo-etal-2025-care, havaldar-etal-2025-culturally, liu-etal-2025-cultural, xu-etal-2025-self}.
For instance, NileChat \cite{el-mekki-etal-2025-nilechat} proposes region-specific models tailored to local communities, while CuMA \cite{sun-etal-2026-cuma} conditions generation on demographic profiles.
However, these methods treat culture as an oversimplified, static attribute or a deterministic categorical label, which fails to reflect modern multicultural realities. 
%
Influenced by globalization and diverse interactional experiences, individuals are not necessarily shaped by a single homogeneous cultural background. 
Instead, their communicative preferences may reflect a composite repertoire~\cite{coste2009plurilingual} formed through exposure to multiple cultural norms, languages, and communities.
Moreover, even when individuals have relatively stable cultural experiences, their situational cultural alignment, namely the communicative strategies they lean toward in a specific context, can shift dynamically throughout a conversation.
Depending on the interlocutor, goal, power relation, and emotional state, speakers may dynamically adapt their communication styles and cultural orientations.
This view aligns with Communication Accommodation Theory \cite{giles1991contexts}, which suggests that people adapt their communicative behavior to evolving social contexts.

To address this limitation, in this paper, we propose \textbf{DyCAC}, a training-free framework that achieves fluid social alignment by integrating \underline{Dy}namic \underline{C}ultural \underline{A}daptation with continuous \underline{C}ognitive tracking.
Unlike prior approaches that treat culture as a static identity constraint, our framework formulates cultural adaptation as a continuous process of belief revision.
Specifically, a Perception module first extracts objective facts, mental-state signals, and cultural cues from the ongoing dialogue.
Building upon this shared evidence, DyCAC performs two complementary processes.
First, it maps cultural cues to a dynamically maintained set of diverse cultural reference profiles and updates their relevance weights across dialogue turns.
Rather than predicting a fixed cultural identity, this dynamic representation is used to derive a soft cultural profile that reflects the interlocutor's potentially hybrid and evolving communicative preferences.
DyCAC further combines this reference-based mixture with directly observed dialogue-style signals, enabling it to capture both composite cultural influences and turn-level shifts in cultural expression.
Second, DyCAC maintains a ToM-Driven Memory architecture that continuously tracks the interlocutor's evolving epistemic states, including beliefs, desires, and intentions.
Finally, the inferred cultural calibration state and cognitive-state representation jointly guide a planning-and-execution module to generate situated, goal-directed responses.
Extensive experiments on social and cultural benchmarks validate the effectiveness of our framework.
It demonstrates enhanced social intelligence in interactive dialogue scenarios and robust generalization across diverse cultural contexts, outperforming other training-free and training-based baselines.

\section{Methodology}

In this section, we detail our proposed DyCAC framework, which decomposes social dialogue generation into four interconnected modules.
We present an overview of the framework in Figure \ref{fig:fig1}.
More details are provided in Appendix~\ref{sec:appendix_details_of_methodology}.

\subsection{Perception}
Our framework first employs a Perception module to extract compact representations of social and contextual cues from the ongoing dialogue. 
Given the current interaction input $x_t$ and the past memory $m_{t-1}$, the module produces a perceptual state:
\begin{equation}
    p_t = \Phi_{\mathrm{perc}}(x_t, m_{t-1}).
\end{equation}
The state $p_t$ contains three categories of evidence: (1) objective facts capturing explicit situations and events; (2) mental-state signals reflecting the immediate intent, affective states, and salient values of the interlocutor; and (3) cultural cues summarizing observable communication styles and social tendencies.
Together, these variables provide the shared evidence required for both cultural adaptation and ToM-driven memory updating.

\subsection{Cultural Profile Inference}
\label{sec:cultural_hypothesis}
Culture influences how people interpret social cues and manage uncertainty in interactions \cite{GUDYKUNST1998227}.
To capture the fluidity of real-world interactions, we conceptualize cultural adaptation not as a static identity classification task, but as a dynamic strategy for reducing interactional friction.

\begin{algorithm}[t]
\caption{Dynamic Cultural Adaptation}
\label{alg:cultural_inference}
\noindent\textbf{Input:} Global cultural pool $\mathcal{H}$, perceptual state $p_t$, current utterance $x_t$, previous cultural profile $\mathbf{c}_{t-1}$, previous culture space $\mathcal{H}_{t-1}$ \\
\noindent\textbf{Output:} Final cultural profile $\mathbf{c}_t$, updated culture space $\mathcal{H}_t$ \\
\noindent\textbf{Notation:} $\ell_{t,i}$: compatibility score, $\pi_{t,i}$: cultural prior

\begin{algorithmic}[1]
\If{$t = 1$}
    \State $\mathcal{H}_1 \gets \mathrm{InitTopK}(p_1, \mathcal{H}, K)$
\Else
    \State $\tilde{\mathcal{H}}_t \gets \mathrm{AugmentToK}(\{\mathbf{h}_{t-1,i} \in \mathcal{H}_{t-1} \mid \Phi_{\mathrm{comp}}(p_t, x_t, \mathbf{h}_{t-1,i}) \ge \theta_{\mathrm{lik}}\}, p_t, x_t, K)$
\EndIf
\For{$\mathbf{h}_{t,i} \in \tilde{\mathcal{H}}_t$}
    \State $\ell_{t,i} \gets \Phi_{\mathrm{comp}}(p_t, x_t, \mathbf{h}_{t,i})$,
    \State $\pi_{t,i} \gets \mathrm{ComputePrior}(p_t, \mathbf{c}_{t-1}, \mathbf{h}_{t,i})$
    \State $\tilde{q}_{t,i} \gets \frac{\pi_{t,i} \cdot \ell_{t,i}}{\sum_{j} \pi_{t,j} \cdot \ell_{t,j}}$
\EndFor
\State $\mathcal{H}_t \gets \{\mathbf{h}_{t,i} \in \tilde{\mathcal{H}}_t \mid \tilde{q}_{t,i} \ge \theta_{\mathrm{filter}}\}$
\State $q_{t,i} \gets \frac{\pi_{t,i} \cdot \ell_{t,i}}{\sum_{j} \pi_{t,j} \cdot \ell_{t,j}}, \quad \forall \mathbf{h}_{t,i} \in \mathcal{H}_t$
\State $\bar{\mathbf{c}}_t \gets \sum_{\mathbf{h}_{t,i} \in \mathcal{H}_{t}} q_{t,i} \cdot \mathbf{h}_{t,i}$
\State $\hat{\mathbf{c}}_t \gets \Phi_{\mathrm{style}}(p_t, x_t)$
\State $\mathbf{c}_t \gets (1-\lambda)\bar{\mathbf{c}}_t + \lambda\hat{\mathbf{c}}_t$
\end{algorithmic}
\end{algorithm}

\boldpara{Global Cultural Reference Pool.}
As shown in Algorithm~\ref{alg:cultural_inference}, we represent culture as a continuous latent space grounded in Hofstede's cultural dimensions \cite{lonner1980culture, hofstede2011dimensionalizing}.
We first define a global cultural reference space $\mathcal{H}$ derived from the Hofstede dimension database\footnote{\url{https://geerthofstede.com/research-and-vsm/}}, formalized as $\mathcal{H} = \{\mathbf{h}_{1}, \mathbf{h}_{2}, \dots, \mathbf{h}_{N}\}$, where $N$ denotes the number of country/region reference profiles.
Each reference profile $\mathbf{h}_{i} \in \mathcal{H}$ is associated with a six-dimensional vector, representing power distance, individualism, masculinity, uncertainty avoidance, long-term orientation, and indulgence.

\boldpara{Turn-Specific Cultural Pool Updating.}
At the first dialogue turn, given the perceptual state $p_1 = \Phi_{\mathrm{perc}}(x_1).$, we leverage LLMs to select the top-$K$ cultural reference anchors whose communication patterns are most compatible with the observed dialogue evidence. 
We then extract corresponding $\mathbf{h}_{i}$ from $\mathcal{H}$ and form an initial culture space $\mathcal{H}_1$:
\begin{equation}
    \mathcal{H}_1 = \{\mathbf{h}_{1,1}, \mathbf{h}_{1,2}, \dots, \mathbf{h}_{1,K}\} \subset \mathcal{H}
\end{equation}
As the dialogue unfolds, rather than committing to a definitive cultural label, we maintain an active culture space $\mathcal{H}_t$ by continuously pruning low-weight cultural profiles and augmenting new profiles, maintaining a constant size of $K$.
Formally, for each previously retained profile $\mathbf{h}_{t-1,i} \in \mathcal{H}_{t-1}$, we compute an utterance-level compatibility score:
\begin{equation}
    \ell_{t,i}
    =
    \Phi_{\mathrm{comp}}(p_t, x_t, \mathbf{h}_{t-1,i}),
\end{equation}%
where $\ell_{t,i}$ measures how likely the utterance $x_t$ is to be produced by someone from culture $\mathbf{h}_{t-1, i}$.
Profiles satisfying $\ell_{t,i} < \theta_{\mathrm{lik}}$ are discarded.
To restore the culture space to exactly $K$ candidates, we prompt the LLM with the current perception $p_t$ and utterance $x_t$ to propose new reference profiles and compute their compatibility score $\ell_{t}$ as well.
This process results in an updated culture space $\tilde{\mathcal{H}}_t$.

In addition to $\ell_{t}$, we calculate \textbf{cultural prior $\pi_{t,i}$} for each candidate profile $\mathbf{h}_{t,i} \in \tilde{\mathcal{H}}_t$.
$\pi_{t,i}$ estimates the probability of the interlocutor originating from culture $\mathbf{h}_{t-1, i}$, given the perceptual state $p_t$ and the previous cultural profile $\mathbf{c}_{t-1}$.
Consequently, we compute the normalized weight $\tilde{q}_{t,i}$ for each profile $\mathbf{h}_{t,i} \in \tilde{\mathcal{H}}_t$, formulated as:
\begin{equation}
\label{eq:normalization}
    \tilde{q}_{t,i} = \frac{\pi_{t,i}  \cdot \ell_{t,i}}{\sum_{\mathbf{h}_{t,j} \in \tilde{\mathcal{H}}_t} \pi_{t,j} \cdot \ell_{t,j}}
\end{equation}
Following this, we further discard cultural profiles whose weight falls below $\theta_{\mathrm{filter}}$ and obtain the final culture space $\mathcal{H}_t$.
\begin{equation}
    \mathcal{H}_t
    =
    \left\{
        \mathbf{h}_{t,i}
        \in
        \tilde{\mathcal{H}}_t
        \mid
        \tilde{q}_{t,i}
        \geq
        \theta_{\mathrm{filter}}
    \right\}.
\end{equation}
We then re-normalize the weights for the remaining profiles in $\mathcal{H}_t$ following the same formulation as Equation~\ref{eq:normalization} to obtain the final weight $q_{t,i}$.

\boldpara{Soft Cultural Calibration.}
Based on these, we formulate the inferred cultural profile $\bar{\mathbf{c}}_t$ as the weighted average of the retained references in $\mathcal{H}_t$:
\begin{equation}
    \bar{\mathbf{c}}_t = \sum_{\mathbf{h}_{t,i} \in \mathcal{H}_{t}} q_{t,i} \cdot \mathbf{h}_{t,i}.
\end{equation}
Although $\mathcal{H}$ is a discrete reference pool, the weighted combination $\bar{\mathbf{c}}_t$ provides a continuous calibration representation that can capture composite cultural influences.
However, country/region-level reference profiles alone may be too coarse to characterize an individual's immediate communicative behavior.
To reduce over-reliance on such population-level references, we additionally infer a direct style from the current interaction:
\begin{equation}
\hat{\mathbf{c}}_t = \Phi_{\mathrm{style}}(p_t, x_t).
\end{equation}
The final cultural profile is then formulated as a blended representation:
\begin{equation}
\mathbf{c}_t = (1-\lambda)\bar{\mathbf{c}}_t + \lambda\hat{\mathbf{c}}_t.
\end{equation}
Through these steps, for example, if an utterance emphasizes personal autonomy and direct decision-making, the module can dynamically shift the weight mass toward cultural profiles characterized by relatively high individualism and low power distance, such as the United States profile.

\subsection{ToM-Driven Memory}
Parallel to cultural inference, the Memory module maintains the continuous evolution of the epistemic state.
It integrates observational evidence with ToM reasoning to track the latent cognitive dynamics of the interlocutor.
At turn $t-1$, the memory state $m_{t-1}$ stores three types of information:
(1) \textbf{observed evidence}, including explicit facts, stated preferences, and interactional events extracted from the perceptual state $p_{t-1}$;
(2) \textbf{inferred mental-state estimates} $z_{t-1}$, including the interlocutor's potential beliefs, desires, and intentions; and
(3) \textbf{dialogue metadata}, such as speaker roles and turn indices.
%
A ToM reasoner first infers latent social variables $z_t$ based on the historical memory $m_{t-1}$ and current perception $p_t$:
\begin{equation}
    z_t = \Phi_{\mathrm{ToM}}(m_{t-1}, p_t).
\end{equation}

Given the context $e_t = \{m_{t-1}, z_t, p_t\}$, an update function $\Phi_{\mathrm{upd}}$ first yields a set of candidate operations $O_{t} = \{o_{t,1}, o_{t,2}, \cdots\}$ for subsequent memory updating.
These operations include: (1) \textbf{assert} novel information; (2) \textbf{revise} existing assumptions; and (3) \textbf{retract} outdated or contradicted data. 

For each candidate operation $o_{t,j} \in$ \{\text{assert}, \text{revise}, \text{retract}\}, a confidence score $ \tau_{t,j} \in (0,1)$ is estimated by the LLM, which indicates the reliability based on the evidence within $e_t$.
The memory $m_t$ is then updated by applying operations whose confidence $\tau_{t,j}$ exceeds a threshold $\tau_{thre}$:
\begin{equation}
m_t = \Phi_{\mathrm{upd}}(e_t, \{o_{t,j} \in O_{t} \mid \tau_{t,j} > \tau_{thre}\}).
\end{equation}
This mechanism allows the system to flexibly execute multiple operations within a single dialogue turn.
Collectively, these epistemic dynamics empower the framework to sustain conversational coherence while accurately tracking the fluid cognitive states of the interlocutor over time.

\subsection{Planning and Execution}
The final module converts the inferred socio-cultural landscape into a situated conversational action.
Given the updated memory $m_t$, the cultural profile $\mathbf{c}_t$, the current interaction input $x_t$, and the designated agent role $r$, the planner formulates a concise action schema:
\begin{equation}
    a_t = \Pi_{\mathrm{plan}}(m_t, \mathbf{c}_t, x_t, r).
\end{equation}
This schema $a_t$ prescribes the strategic function of the upcoming turn, dictating behaviors such as probing, aligning, conceding, or establishing boundaries.
Ultimately, the executor operationalizes this schema into the final response:
\begin{equation}
    y_t = \Pi_{\mathrm{exec}}(x_t, a_t, \mathbf{c}_t, r).
\end{equation}
Within this framework, the planning phase dictates what objective the utterance must achieve, while the execution phase determines how that goal is expressed under the guidance of the cultural profile.

\begin{table*}[htbp]
\centering

\footnotesize

\renewcommand{\arraystretch}{0.98}

\resizebox{\textwidth}{!}{
\begin{tabular}{l l c c c c c c c c}
\toprule

\multirow{2}{*}{\textbf{Model}} 
& \multirow{2}{*}{\textbf{Method}} 
& \multirow{2}{*}{\shortstack{\textbf{\textsc{Bel}}\\ \textbf{[0, 10]}}}
& \multirow{2}{*}{\shortstack{\textbf{\textsc{Rel}}\\ \textbf{[-5, 5]}}}
& \multirow{2}{*}{\shortstack{\textbf{\textsc{Kno}}\\ \textbf{[0, 10]}}}
& \multirow{2}{*}{\shortstack{\textbf{\textsc{Sec}}\\ \textbf{[-10, 0]}}}
& \multirow{2}{*}{\shortstack{\textbf{\textsc{Soc}}\\ \textbf{[-10, 0]}}}
& \multirow{2}{*}{\shortstack{\textbf{\textsc{Fin}}\\ \textbf{[-5, 5]}}}
& \multirow{2}{*}{\shortstack{\textbf{\textsc{Goal}}\\ \textbf{[0, 10]}}}
& \multirow{2}{*}{\textbf{Overall}\hspace{0.6em}} \\

\\

\midrule

\multirow{9}{*}{\makebox[2.6em][c]{\rotatebox[origin=c]{90}{Qwen2.5-7B}}}
& Vanilla & 8.73 & 3.24 & 5.65 & -0.34 & -0.14 & 0.41 & 7.39 & 3.56 \\
& CoT & 8.68 & 3.16 & 5.90 & -0.37 & -0.16 & 0.47 & 7.48 & 3.59 \\
& ReAct & 8.76 & 3.19 & 5.76 & -0.31 & -0.09 & 0.41 & 7.40 & 3.59 \\
& MAD & 8.57 & 3.23 & 5.32 & -0.23 & -0.10 & 0.56 & 7.61 & 3.57 \\
& MetaMind & 8.79 & \textbf{3.60} & 5.93 & \textbf{-0.12} & -0.09 & 0.48 & 7.45 & 3.72 \\
& SDPO & 8.66 & 3.50 & 5.38 & -0.15 & \textbf{-0.07} & \textbf{0.63} & 7.43 & 3.63 \\
& Sotopia-$\Omega$ & \textbf{8.98} & 3.38 & 5.88 & -0.16 & -0.11 & 0.51 & 7.39 & 3.70 \\
& Sotopia-RL & 8.84 & 3.52 & 5.85 & -0.31 & -0.15 & 0.58 & 7.91 & 3.75 \\
& \cellcolor{purple!10}\textbf{DyCAC} & \cellcolor{purple!10}8.89 & \cellcolor{purple!10}3.33 & \cellcolor{purple!10}\textbf{6.11}\rlap{$^{**}$} & \cellcolor{purple!10}-0.16 & \cellcolor{purple!10}-0.12 & \cellcolor{purple!10}0.60 & \cellcolor{purple!10}\textbf{7.93}\rlap{$^{**}$} & \cellcolor{purple!10}\textbf{3.80}\rlap{$^{**}$} \\

\midrule

\multirow{6}{*}{\makebox[2.6em][c]{\rotatebox[origin=c]{90}{Gemma4-26B}}}
& Vanilla & 9.16 & 3.23 & 5.77 & -0.60 & -0.21 & 0.61 & 7.76 & 3.67 \\
& CoT & \textbf{9.25} & 3.20 & 5.82 & -0.38 & -0.20 & 0.64 & 7.69 & 3.72 \\
& ReAct & 9.01 & 3.14 & 5.41 & -0.34 & -0.19 & 0.67 & 7.49 & 3.60 \\
& MAD & 9.14 & 3.16 & 5.17 & -0.39 & \textbf{-0.18} & 0.72 & 7.69 & 3.62 \\
& MetaMind & 9.01 & \textbf{3.27} & 5.84 & \textbf{-0.13} & -0.22 & 0.58 & 7.54 & 3.70 \\
& \cellcolor{purple!10}\textbf{DyCAC} & \cellcolor{purple!10}9.22 & \cellcolor{purple!10}3.20 & \cellcolor{purple!10}\textbf{6.15}\rlap{$^{*}$} & \cellcolor{purple!10}-0.22\rlap{$^{*}$} & \cellcolor{purple!10}-0.23 & \cellcolor{purple!10}\textbf{0.75} & \cellcolor{purple!10}\textbf{7.88} & \cellcolor{purple!10}\textbf{3.82}\rlap{$^{*}$} \\

\bottomrule
\end{tabular}
}
\caption{Evaluation results on the SOTOPIA benchmark. Each experiment is run three times, and the reported results are averaged. The best performance is bolded.
$^{*}p<0.05$ and $^{**}p<0.01$ indicate statistically significant differences compared with the vanilla method.}
\label{tab:results_on_sotopia}
\vspace{-1em}
\end{table*}

\section{Experiments}

\subsection{Experimental Settings}

\boldpara{Baselines.}
Our proposed DyCAC framework is a training-free framework based on the integration of dynamic cultural adaptation and cognitive state tracking.
For a fair comparison, we mainly compare it to the following training-free baselines: \textbf{Chain of Thought} (CoT; \citet{wei-2022-cot}), \textbf{ReAct} \cite{yao2023react}, \textbf{Multi-Agent Debate} (MAD; \citet{liang-etal-2024-encouraging}), \textbf{SocialGaze} \cite{vijjini-etal-2024-socialgaze}, \textbf{SocialAgent} \cite{yuan-etal-2024-measuring}, \textbf{CulturalDebate} \cite{ki-etal-2025-multiple}, and \textbf{MetaMind} \cite{zhang2026metamind}.
Moreover, a comparison is conducted among several training-based methods, including \textbf{SDPO} \cite{kong-etal-2025-sdpo}, \textbf{SOTOPIA-$\Omega$} \cite{zhang-etal-2025-sotopia}, and \textbf{SOTOPIA-RL} \cite{yu2025sotopiarl}.
More baseline details about baseline methods are described in Appendix \ref{sec:appendix_details_of_baselines}.

\boldpara{Benchmarks.}
We evaluate the proposed framework on two benchmarks designed to assess social intelligence and culturally grounded affective understanding, respectively. 
Further benchmark details are provided in Appendix \ref{sec:appendix_details_of_benchmarks}.
\begin{itemize}[leftmargin=*]
    \item \textbf{SOTOPIA} \cite{zhou2024sotopia} is an open-ended interactive environment for evaluating social intelligence.
    It features 90 social scenarios, 40 agents, and 90 relationships.
    For each scenario, five pairs of characters are sampled, resulting in 450 tasks.
    We report performance across seven dimensions: Believability \textsc{(Bel)}, Relationship \textsc{(Rel)}, Knowledge \textsc{(Kno)}, Secret \textsc{(Sec)}, Social Rules \textsc{(Soc)}, Financial and Material Benefits \textsc{(Fin)}, and Goal Completion \textsc{(Goal)}.
    \item \textbf{CEDAR} \cite{dai-etal-2026-tears} is a multilingual benchmark evaluating culture-specific emotion alignment.
    It covers seven languages and 14 fine-grained emotion categories.
    We utilize both its multimodal (400 instances per language) and text-only (1,166 instances per language) subsets to evaluate culturally situated understanding.
\end{itemize}

\boldpara{Implementation Details.}
In this study, we employ four foundation models: \texttt{Qwen2.5-7B} \cite{qwen2.5}, \texttt{Gemma3-12B} \cite{gemma_2025}, \texttt{Gemma4-26B}\footnote{\url{https://huggingface.co/google/gemma-4-26B-A4B-it}}, and \texttt{Qwen3.5-35B} \cite{qwen3.5}.
All models are deployed via vLLM \cite{10.1145/3600006.3613165}.
For the SOTOPIA evaluation, we adopt \texttt{GPT-4o} (version: 2024-11-20; \citet{openai2024gpt4ocard}) as the evaluator and \texttt{Gemini2.5-Flash} \cite{comanici2025gemini25pushingfrontier} as the interaction companion agent.
All experiments are conducted on two RTX A6000 GPUs with the temperature set to 0 during inference.

\begin{table*}[t]
\centering
\footnotesize
\renewcommand{\arraystretch}{1.05}
\setlength{\tabcolsep}{2.2pt}

\resizebox{\textwidth}{!}{
\begin{tabular}{@{}l l| c c c c c c c c| c c c c c c c c@{}}
\toprule

\multirow{2}{*}{\textbf{Model}} 
& \multirow{2}{*}{\textbf{Method}} 
& \multicolumn{8}{c|}{\textbf{Multimodal}}
& \multicolumn{8}{c}{\textbf{Text-Only}} \\

\cmidrule(lr){3-10} \cmidrule(lr){11-18}

&
& \textbf{AR}
& \textbf{EN}
& \textbf{ES}
& \textbf{HI}
& \textbf{JA}
& \textbf{SW}
& \textbf{ZH}
& \textbf{Avg.}
& \textbf{AR}
& \textbf{EN}
& \textbf{ES}
& \textbf{HI}
& \textbf{JA}
& \textbf{SW}
& \textbf{ZH}
& \textbf{Avg.} \\

\midrule

\multirow{6}{*}{\makebox[3.2em][c]{\rotatebox[origin=c]{90}{Gemma3-12B}}}
& Vanilla 
& 34.51 & 34.51 & 40.05 & 27.13 & 28.57 & 30.15 & 15.04 & 29.99
& 37.42 & 34.51 & 44.80 & 23.73 & 33.19 & 34.91 & 11.23 & 31.40 \\

& SocialGaze 
& 34.38 & 38.66 & 41.03 & 36.73 & 33.33 & 30.83 & 29.89 & 34.98
& 42.52 & 49.73 & \best{46.19} & 33.82 & 30.50 & 33.25 & 31.19 & 38.17 \\

& SocialAgent 
& 34.76 & 34.26 & 34.59 & 33.50 & \best{34.50} & \best{33.50} & 27.14 & 33.18
& 41.17 & 49.49 & 39.54 & 42.02 & 37.39 & 36.54 & 33.19 & 39.91 \\

& CulturalDebate 
& \best{38.99} & 39.14 & \best{43.80} & 38.30 & 32.83 & 32.50 & 28.39 & 36.28
& \best{43.85} & \best{52.46} & 44.63 & 42.36 & 35.36 & 38.08 & 32.73 & 41.35 \\

& MetaMind 
& 32.57 & 36.03 & 42.17 & 36.65 & 33.07 & 29.26 & \best{31.70} & 34.49
& 38.33 & 47.36 & 38.16 & 40.23 & 35.37 & 36.27 & 33.56 & 38.47 \\

& DyCAC
& 35.37 & \best{39.55} & 42.76 & \best{46.88} & 34.18 & \best{33.50} & 30.62 & \best{37.55}
& 42.95 & 50.92 & 44.54 & \best{44.20} & \best{38.90} & \best{38.21} & \best{33.59} & \best{41.90} \\

\midrule

\multirow{6}{*}{\makebox[3.2em][c]{\rotatebox[origin=c]{90}{Qwen3.5-35B}}}
& Vanilla 
& 26.65 & 37.82 & 30.00 & 37.00 & 29.83 & 32.25 & 26.96 & 31.50
& 40.07 & 43.57 & 43.82 & 39.84 & 34.36 & 32.22 & 25.93 & 37.12 \\

& SocialGaze 
& 37.94 & 39.10 & 35.31 & 38.02 & 26.82 & 32.08 & 28.89 & 34.02
& 42.99 & 47.33 & 44.24 & 39.59 & 34.05 & 36.08 & 27.64 & 38.85 \\

& SocialAgent 
& \best{39.55} & 42.96 & 37.34 & 38.79 & \best{33.83} & 30.08 & 33.67 & 36.60
& \best{43.55} & 46.84 & 40.79 & \best{43.79} & 36.18 & 34.04 & 39.86 & 40.72 \\

& CulturalDebate
& 37.69 & 41.32 & 39.23 & 40.12 & 32.25 & 29.97 & 38.11 & 36.96
& 42.96 & \best{47.98} & 43.95 & 42.49 & 34.22 & 35.25 & 43.86 & 41.53 \\

& MetaMind 
& 39.39 & 38.52 & \best{40.00} & 36.01 & 33.76 & 30.98 & 32.65 & 35.90
& 41.17 & 43.81 & 36.39 & 39.93 & 35.08 & 29.33 & 36.72 & 37.49 \\

& DyCAC
& 36.00 & \best{43.63} & 39.03 & \best{42.41} & 30.15 & \best{32.18} & \best{38.50} & \best{37.41}
& 42.08 & 46.78 & \best{44.50} & 41.55 & \best{36.71} & \best{36.96} & \best{44.16} & \best{41.82} \\

\bottomrule

\end{tabular}
}

\caption{
Results on the CEDAR benchmark. The best performance in each column is highlighted. Languages: AR = Arabic, EN = English, ES = Spanish, HI = Hindi, JA = Japanese, SW = Swahili, ZH = Chinese.
}
\label{tab:cedar_results}

\end{table*}
\begin{table*}[t]
\centering
\small 
\renewcommand{\arraystretch}{1.0} 

\resizebox{\linewidth}{!}{
\begin{tabular}{@{} l c c c c c c c c @{}} 
\toprule
\multirow{2}{*}{\textbf{Method}} & \textbf{\textsc{Bel}} & \textbf{\textsc{Rel}} & \textbf{\textsc{Kno}} & \textbf{\textsc{Sec}} & \textbf{\textsc{Soc}} & \textbf{\textsc{Fin}} & \textbf{\textsc{Goal}} & \multirow{2}{*}{\textbf{Overall}} \\
 & [0, 10] & [-5, 5] & [0, 10] & [-10, 0] & [-10, 0] & [-5, 5] & [0, 10] & \\
\midrule
Vanilla & 8.73 & 3.24 & 5.65 & -0.34 & -0.14 & 0.41 & 7.39 & 3.56 \\
\textbf{DyCAC (Full)} & \textbf{8.89} & \textbf{3.33} & \textbf{6.11} & \textbf{-0.16} & \textbf{-0.12} & \textbf{0.60} & \textbf{7.93} & \textbf{3.80} \\
\midrule

\rowcolor{gray!30} 
\multicolumn{9}{l}{\textit{\textbf{Ablation Study on Framework Modules}}} \\
\hspace{1em}\textit{w/o} Perception & 8.75 & 3.16 & 5.60\rlap{$^{**}$} & -0.16 & -0.13 & 0.53 & 7.74 & 3.64 \\
\hspace{1em}\textit{w/o} Cultural Profile Inference & 8.72 & 2.98\rlap{$^{*}$} & 5.69\rlap{$^{*}$} & -0.15 & -0.15 & 0.41 & 7.77 & 3.61 \\
\hspace{1em}\textit{w/o} Full Memory Module & 8.57\rlap{$^{**}$} & 3.23 & 5.90 & -0.22 & -0.11 & 0.58 & 7.58\rlap{$^{*}$} & 3.65 \\
\midrule

\rowcolor{gray!30} 
\multicolumn{9}{l}{\textit{\textbf{Ablation Study on ToM Mechanism}}} \\
\hspace{1em}\textit{w/o} Latent ToM Tracking & 8.72 & 3.20 & 5.93 & -0.25 & -0.12 & 0.48 & 7.64 & 3.66 \\
\bottomrule
\end{tabular}
}
\caption{
Ablation of the framework modules and ToM mechanism on the SOTOPIA benchmark using Qwen2.5-7B.
$^{*}p<0.05$ and $^{**}p<0.01$ denote statistically significant performance drops compared to the full framework.
}
\label{tab:ablation_study}
\vspace{-1em}
\end{table*}

\subsection{Main Results}

\boldpara{Results on SOTOPIA.}
We present the results on the SOTOPIA benchmark in Table~\ref{tab:results_on_sotopia}.
Across both backbones, our framework achieves the highest overall performance, surpassing both training-free and training-based methods.
Beyond this, individual metrics highlight varying abilities among the baselines.
While MetaMind excels in Relationship (\textsc{Rel}) and Secret (\textsc{Sec}) by explicitly modeling cognitive reasoning, Sotopia-$\Omega$ maximizes Believability (\textsc{Bel}) through injecting multi-step negotiation strategies via behavior cloning.
Crucially, our approach demonstrates pronounced dominance in Knowledge (\textsc{Kno}) and Goal Completion (\textsc{Goal}).
These gains can be attributed to our core mechanisms.
First, the ToM-Driven Memory enhances knowledge by actively modeling the interlocutor's hidden states to strategically resolve information asymmetries.
Second, by grounding dynamic cultural calibration in these ToM-derived insights, the framework crafts culturally aligned proposals that minimize interactional friction and foster cooperation.
This adaptive approach also enables the framework to effectively secure its objectives, yielding the superior Goal Completion and Financial (\textsc{Fin}) scores observed with Gemma4-26B.

\boldpara{Results on CEDAR.}
We further evaluate the framework on the CEDAR benchmark, with results shown in Table \ref{tab:cedar_results}.
Our framework achieves the highest average performance across both backbones, demonstrating exceptional generalization across diverse language contexts.
Moreover, the performance breakdown reveals two insights.
First, our approach exhibits the most pronounced improvements in languages such as Hindi, Swahili, and Chinese.
This suggests that our dynamic cultural adaptation effectively mitigates the inherent geographical biases of language models.
Unlike CulturalDebate that relies on multi-agent debate to reach a cultural consensus, our framework maps observable cues into a continuous mixture to calibrate the implicit emotional nuances prevalent in these cultures.
Second, our framework outperforms advanced cognitive approaches such as MetaMind. 
While existing methods excel in tracking mental states, emotions and hidden intentions manifest differently across cultures. 
Thus, grounding cognitive inference in dynamic cultural calibration facilitates the context-sensitive interpretation of ambiguous affective cues.

\begin{figure*}[t]
\centering
\includegraphics[width=\textwidth]{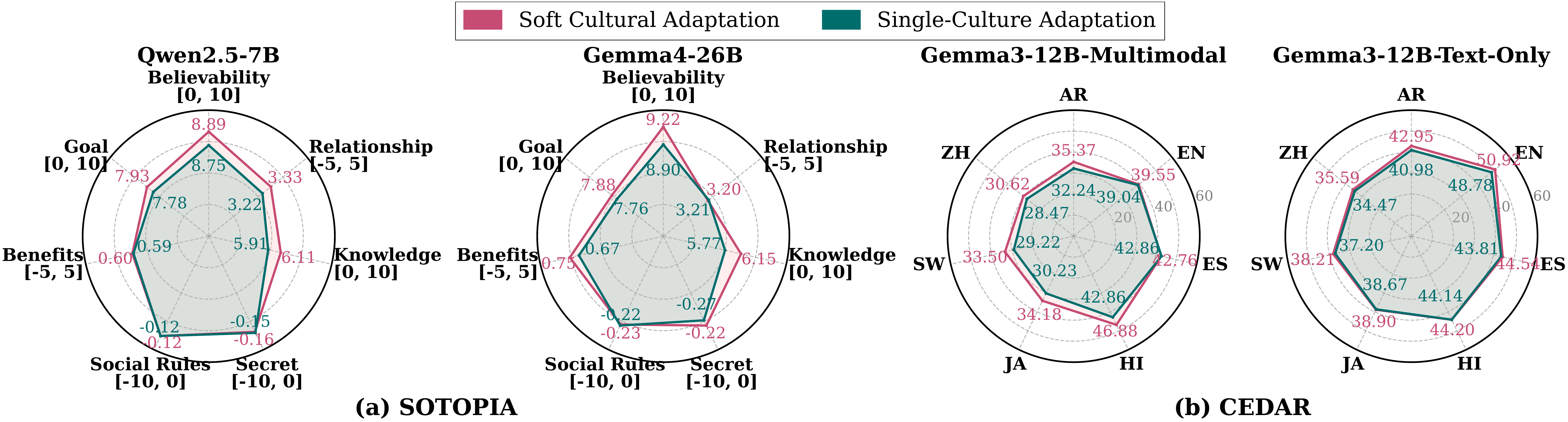} 
\caption{
Performance comparison between Soft Cultural Adaptation (SCA) and Single-Culture Adaptation (SiCA) on the SOTOPIA and CEDAR benchmarks.
}
\label{fig:radar_cultural_adaptation}
\vspace{-1em}
\end{figure*}

\subsection{Method Analysis}

\boldpara{Ablation Study.}
To evaluate the contribution of each core component, we conduct an ablation study on the SOTOPIA benchmark, with results presented in Table~\ref{tab:ablation_study}.
First, removing the Perception module leads to a notable degradation in Knowledge, as without extracting key information from the dialogue, downstream modules are deprived of the evidence necessary to utilize factual context.
Similarly, discarding Cultural Profile Inference impairs Relationship, Knowledge, and Financial Benefits.
This indicates that lacking dynamic cultural calibration causes the framework to misalign with the interlocutor and misinterpret social cues when deploying knowledge.
Crucially, ablating the Full Memory Module causes a pronounced decline in Believability and Goal Completion.
Without a persistent record of the interaction and the interlocutor's cognitive dynamics, the framework struggles to maintain conversational continuity, act naturally, and advance strategic objectives.
To further isolate the impact of mental state modeling, we disable latent ToM tracking within the Memory Module, while preserving explicit memory operations and updates.
The resulting performance decay reveals that without dynamically inferring hidden desires and shifting intentions, merely recording dialogue history is insufficient for social alignment.

\begin{table}[t]
\centering
\footnotesize
\renewcommand{\arraystretch}{0.98}
\setlength{\tabcolsep}{8pt}

\resizebox{0.8\columnwidth}{!}{%
\begin{tabular}{@{}c l c c@{}}
\toprule

\textbf{Model} & \textbf{Method} & \multicolumn{2}{c}{\textbf{Avg.}} \\

\midrule

\multirow{5}{*}{\makebox[2.5em][c]{\rotatebox[origin=c]{90}{\scriptsize Gemma3-12B}}}
& Vanilla & 29.99 & \underline{31.40} \\
& First-Person Role-Play & 29.72 & \underline{31.66} \\
& Third-Person Role-Play & 33.35 & \underline{34.75} \\
& SiCA & 34.99 & \underline{41.15} \\
& SCA (Ours) & \textbf{37.55} & \underline{\textbf{41.90}} \\

\midrule

\multirow{5}{*}{\makebox[2.5em][c]{\rotatebox[origin=c]{90}{\scriptsize Qwen3.5-35B}}}
& Vanilla & 31.50 & \underline{37.12} \\
& First-Person Role-Play & 33.67 & \underline{38.43} \\
& Third-Person Role-Play & 34.50 & \underline{39.46} \\
& SiCA & 36.88 & \underline{40.77} \\
& SCA (Ours) & \textbf{37.41} & \underline{\textbf{41.82}} \\

\bottomrule

\end{tabular}%
}

\caption{
Method comparison on the CEDAR benchmark.
The two Avg. scores denote the results on multimodal (left) and text-only (right) subsets, respectively.
}
\label{tab:roleplay_results}
\vspace{-1em}
\end{table}

\boldpara{Effect of Dynamic Cultural Adaptation.}
To validate the dynamic cultural adaptation introduced in \S\ref{sec:cultural_hypothesis}, we compare our Soft Cultural Adaptation (SCA) strategy against a rigid Single-Culture Adaptation (SiCA) baseline.
Specifically, SCA achieves soft cultural calibration by combining an inferred cultural profile with a directly extracted dialogue-style vector.
In contrast, SiCA restricts its adaptation to only the single cultural profile with the highest weight.
As illustrated in Figure \ref{fig:radar_cultural_adaptation}, SCA consistently outperforms SiCA on both the SOTOPIA and CEDAR benchmarks across backbone models.
These performance gaps stem from how each method handles cultural calibration.
While SiCA anchors the model to a single, static cultural prototype, SCA represents culture as a continuous latent space and employs dynamic updating to mitigate this rigidity.
Therefore, SCA transcends rigid stereotypical biases, achieving fluid alignment with the interlocutor's evolving socio-cultural dynamics.

Furthermore, we evaluate our cultural adaptation strategy on the CEDAR benchmark.
We compare our SCA method against three baselines: First-Person Role-Play, Third-Person Role-Play, and SiCA.
As shown in Table~\ref{tab:roleplay_results}, while conventional persona assignments offer marginal improvements over the baseline, they consistently underperform both SiCA and SCA methods.
This highlights a critical limitation of traditional cultural role-playing approaches. 
%
By modeling cultural background as a dynamic mixture rather than a fixed label, our framework effectively captures how cultural nuances shape emotional expression, leading to better cross-cultural emotion understanding.

\begin{figure*}[t]
\centering
\includegraphics[width=0.95\textwidth]{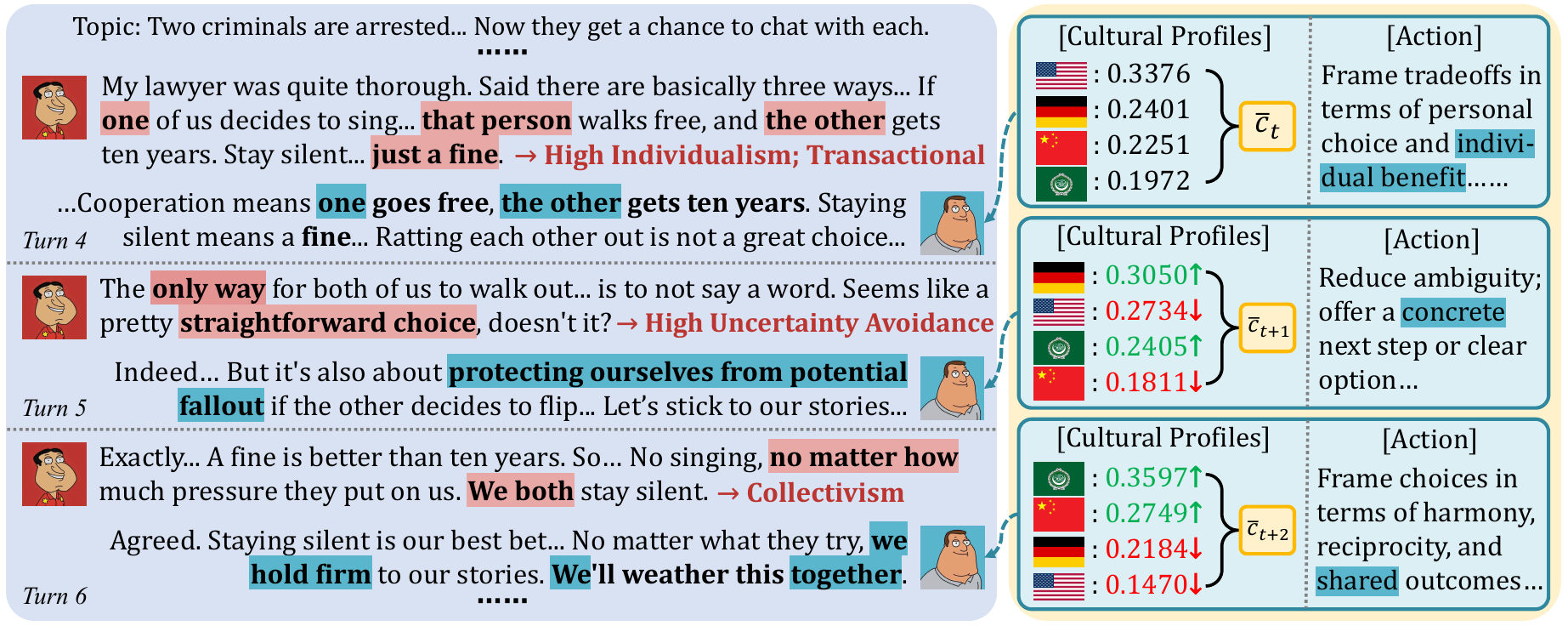} 
\caption{
A qualitative example of dynamic cultural adaptation in our framework.
As the dialogue progresses, our framework captures subtle semantic shifts (Left) to continuously update the latent cultural profile, translating these beliefs into effective strategic actions (Right).
}
\label{fig:case_study}
\vspace{-1em}
\end{figure*}

\subsection{Case Study}
To illustrate the dynamic adaptation process of the DyCAC framework, we analyze a case situated in a Prisoner's Dilemma scenario, as shown in Figure~\ref{fig:case_study}.
%
%
%
Initially, the interlocutor adopts a transactional stance, emphasizing individual consequences (``one goes free'').
Based on this intent, our framework initializes a soft cultural mixture weighted toward high-individualism profiles, aligning with metrics typical of the United States in Hofstede's database \cite{hofstede2011dimensionalizing}.
This guides the framework to adopt an action schema that prioritizes personal benefit.
However, as the dialogue evolves, the interlocutor's latent beliefs and desires shift.
Consequently, the observed cultural expression transitions from individualism to uncertainty avoidance, and later manifests as collectivism. 
Unlike static labeling methods that fail to capture these shifting dynamics, our framework actively tracks the cognitive states and dynamically refines its cultural mixture based on the ongoing conversational evidence.
As a result, our framework adapts its strategy from prioritizing personal benefit to fostering mutual cooperation (``We'll weather this together'').
This demonstrates the effectiveness of integrating soft cultural adaptation with ToM mechanisms for social alignment.

\section{Related Work}

\subsection{Social Dialogue}
Numerous studies have focused on language models for social dialogue \cite{xu-etal-2022-beyond, wang-etal-2024-sotopiapi, jafari-etal-2025-beyond, song2025survivalevaluatingllmssocial, xu-etal-2025-multiagentesc, Hu_Wang_Xie_Li_Ma_Guo_2026, zhang2026metamind, zhang2026intervensiminterventionawaresocialnetwork}, where models are expected to pursue social goals and adapt their strategies over multi-turn interactions.
Early efforts established benchmarks such as SOTOPIA \cite{zhou2024sotopia}, SocialMaze \cite{xu2025socialmazebenchmarkevaluatingsocial} and SocialEval \cite{zhou-etal-2025-socialeval} to evaluate LLMs in socially grounded scenarios.
To improve performance on social tasks, researchers have proposed various training-based approaches, which optimize models using social feedback \cite{wang2026adaptive}, reward signals \cite{kong-etal-2025-sdpo, yu2025sotopiarl}, or structured reasoning strategies \cite{liu-etal-2025-epo, zhang-etal-2025-sotopia, zhang-etal-2026-semantic}.
Concurrently, training-free methods elicit social competence at inference time by employing multiple agents to simulate diverse social perspectives \cite{vijjini-etal-2024-socialgaze, yuan-etal-2024-measuring} or coordinate decisions \cite{bhattacharyya2026social}.
However, these methods typically neglect the critical role of cultural nuances in decoding interpersonal signals, thereby constraining their effectiveness in diverse social environments.

\subsection{Cultural Adaptation}
Recent work has increasingly explored robust methods for cultural adaptation \cite{alkhamissi-etal-2024-investigating, havaldar-etal-2025-culturally, song-etal-2025-injecting, zhang2026evaluating}.
Early efforts primarily leveraged synthesized data to enhance the cultural awareness of LLMs \cite{NEURIPS2024_culturellm, el-mekki-etal-2025-nilechat, guo-etal-2025-care, xu-etal-2025-self}.
For example, CulturePark \cite{NEURIPS2024_culturepark} enhances cultural reasoning by training models on cultural dialogues synthesized through multi-agent interactions.
More recently, some studies have proposed specialized training paradigms for cross-cultural alignment \cite{liu-etal-2025-cultural, sun-etal-2026-cuma}.
For instance, CulFiT \cite{feng-etal-2025-culfit} aligns model behaviors with specific cultural values via culturally grounded instruction tuning.
Moreover, agent-based approaches have also been explored for cultural alignment, including multi-agent collaboration \cite{lica2026mindforge} and debate \cite{ki-etal-2025-multiple}.
However, these paradigms typically treat culture as a static attribute, failing to capture the fluid evolution of cultural expressions.

\section{Conclusion}
In this paper, we present \textbf{DyCAC}, a training-free framework for fluid social alignment that integrates dynamic cultural calibration with continuous cognitive tracking.
DyCAC derives a soft and dynamically updated cultural calibration state from population-level cultural references and observable dialogue-style cues, thereby accommodating composite cultural influences and context-dependent shifts in communicative behavior.
Concurrently, a ToM-driven memory actively tracks the latent cognitive dynamics of the interlocutor.
Experiments on the SOTOPIA and CEDAR benchmarks demonstrate that DyCAC consistently improves over strong baselines in interactive social dialogue and culturally grounded emotion understanding.

\section*{Acknowledgments}
This work was supported in part by the National Science and Technology Major Project of China under Grant 2026ZD0125700 and in part by the National Natural Science Foundation of China under Grant 62402158, and in part by the Key Science \& Technology Project of Anhui Province under Grant 202523j08050001.

\section*{Limitations}
Despite the promising results of the DyCAC framework, our method is subject to several limitations that remain to be addressed in future research.
\boldpara{Dependency on Base Model Capabilities.}
As a training-free framework, DyCAC fundamentally relies on the zero-shot reasoning capabilities of the underlying foundation models (e.g., Qwen2.5-7B). 
Consequently, the accuracy and reliability of its continuous cultural calibration and cognitive tracking are inherently constrained by the overall performance of these models.

\boldpara{Scope of the Cultural Space.}
While we use Hofstede’s dimensions to establish the continuous culture space, this theory fundamentally categorizes culture at the level of entire countries or broad geographic regions.
Consequently, our framework currently captures broad national differences rather than highly specific subcultures or localized subcultures.

\section*{Ethical Considerations}
The development of the DyCAC framework necessitate adherence to ethical standards in cultural modeling.
\boldpara{Nature of the Framework.}
Our proposed framework is designed to enhance social and cultural alignment in language models.
By conceptualizing culture as a dynamic, continuous latent mixture, the framework actively avoids rigid cultural stereotyping.
This ensures the model adapts to the nuanced, hybrid communicative behaviors of individuals in a respectful and non-reductive manner. 
Furthermore, the integration of cognitive tracking is explicitly intended to reduce interactional friction and foster mutual cooperation, rather than to manipulate user behavior.

\bibliography{custom}

\appendix

\section{Details of Methodology}
\label{sec:appendix_details_of_methodology}

\subsection{Hofstede's Cultural Dimensions}
Hofstede's cultural dimensions \cite{hofstede2011dimensionalizing} provide a comparative framework for describing cross-cultural differences in social values and communication expectations.
In this work, we use these dimensions as soft style indicators rather than deterministic identity labels.
Each country or region is represented by a six-dimensional vector whose components correspond to the following dimensions.

\begin{itemize}[leftmargin=*]
    \item \textbf{Power Distance (PDI)} measures the extent to which unequal power relations and hierarchical authority are accepted.
    \item \textbf{Individualism (IDV)} measures the extent to which people emphasize individual autonomy rather than group affiliation.
    \item \textbf{Masculinity (MAS)} measures the extent to which a culture emphasizes competition, achievement, and assertiveness rather than care, cooperation, and quality of life.
    \item \textbf{Uncertainty Avoidance (UAI)} measures the extent to which people prefer rules, predictability, and ambiguity reduction.
    \item \textbf{Long Term Orientation (LTO)} measures the extent to which people value future-oriented persistence, adaptation, and delayed rewards.
    \item \textbf{Indulgence (IVR)} measures the extent to which people allow gratification of desires and enjoyment of life rather than restraining them through strict social norms.
\end{itemize}

\subsection{Detailed Cultural Strategies}
\label{sec:appendix_cultural_strategies}
The definitions of strategies are based on Hofstede's cultural dimensions \cite{hofstede2011dimensionalizing}.
We operationalize the cultural calibration vector $\mathbf{c}_t$ through a small set of culture-conditioned strategy directives, thus providing guidance for planning and execution.
Let \(\mathcal{D}=\{\mathrm{PDI}, \mathrm{IDV}, \mathrm{MAS}, \mathrm{UAI}, \mathrm{LTO}, \mathrm{IVR}\}\) denote the six cultural dimensions.
For each dimension \(d \in \mathcal{D}\), the corresponding score \(\mathbf{c}_t^{(d)} \in [0, 100]\) is discretized into a coarse strategy level:
\begin{equation}
\tau(\mathbf{c}_t^{(d)}) =
\begin{cases}
\textsc{low}, & \mathbf{c}_t^{(d)} < 35,\\
\textsc{medium}, & 35 \leq \mathbf{c}_t^{(d)} \leq 65,\\
\textsc{high}, & \mathbf{c}_t^{(d)} > 65.
\end{cases}
\end{equation}
As shown in Table~\ref{tab:cultural_strategy_directives}, at turn \(t\), the planner retrieves the strategy directive for each pair \((d,\tau(\textbf{c}_t^{(d)}))\).
The retrieved directives are used as soft control signals for the planner and executor.
They do not determine the dialogue act itself; instead, they calibrate how the selected action should be framed (e.g., in terms of directness, uncertainty management, and relational emphasis).

\begin{table*}[t]
\centering
\small
\begin{tabular}{p{0.10\linewidth} p{0.12\linewidth} p{0.70\linewidth}}
\toprule
\textbf{Dimension} & \textbf{Level} & \textbf{Culture-Conditioned Strategy} \\
\midrule
\multirow{3}{*}{PDI}
& \textsc{high}
& Show respect for hierarchy and preserve face; phrase pushes as suggestions rather than commands. \\
& \textsc{medium}
& Be polite and respectful without sounding stiff. \\
& \textsc{low}
& Speak as an equal; be plain, direct, and collaborative. \\
\midrule
\multirow{3}{*}{IDV}
& \textsc{high}
& Frame tradeoffs in terms of personal choice and individual benefit. \\
& \textsc{medium}
& Balance personal benefit with relationship impact. \\
& \textsc{low}
& Frame choices in terms of harmony, reciprocity, and shared outcomes. \\
\midrule
\multirow{3}{*}{MAS}
& \textsc{high}
& Be decisive and outcome-focused, but stay socially smooth. \\
& \textsc{medium}
& Balance empathy with concrete progress. \\
& \textsc{low}
& Prioritize warmth, cooperation, and rapport over dominance. \\
\midrule
\multirow{3}{*}{UAI}
& \textsc{high}
& Reduce ambiguity; offer a concrete next step or clear option. \\
& \textsc{medium}
& Be clear while leaving room for flexibility. \\
& \textsc{low}
& Keep the tone adaptive and conversational; do not over-structure. \\
\midrule
\multirow{3}{*}{LTO}
& \textsc{high}
& Emphasize long-term trust and downstream benefits. \\
& \textsc{medium}
& Balance immediate progress with future consequences. \\
& \textsc{low}
& Focus on the immediate interaction and near-term payoff. \\
\midrule
\multirow{3}{*}{IVR}
& \textsc{high}
& Use a relaxed, warm, and expressive tone when appropriate. \\
& \textsc{medium}
& Keep affect balanced and natural. \\
& \textsc{low}
& Use measured, disciplined language; avoid overdoing emotion. \\
\bottomrule
\end{tabular}
\caption{
Detailed cultural strategy directives.
}
\label{tab:cultural_strategy_directives}
\end{table*}

\begin{figure}[t]
\centering
\includegraphics[width=\columnwidth]{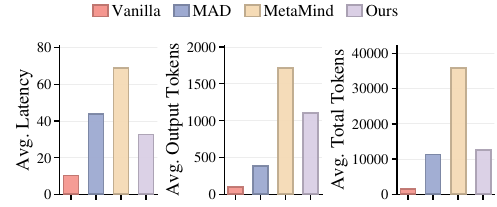} 
\vspace{-1em}
\caption{
Efficiency comparison of our framework and baseline methods on SOTOPIA. We report average latency, average number of output tokens, and average total tokens consumed per dialogue turn.
}
\label{fig:latency_and_token}
\vspace{-1em}
\end{figure}

\section{Details of Experiments}

\subsection{Details of Baselines}
\label{sec:appendix_details_of_baselines}
This section provides details about the baseline methods used in our experiments. 
We mainly compare our framework with training-free prompting and agent-based methods.
\begin{itemize}[leftmargin=*]
    \item \textbf{Chain-of-Thought prompting} (CoT; \citet{wei-2022-cot}) asks a language model to produce intermediate reasoning steps before giving the final answer. We use the prompt template ``Let's think step by step'' in our experiments.

    \item \textbf{ReAct} \cite{yao2023react} is a prompting framework that interleaves reasoning traces with task-specific actions. We set the maximum number of ReAct steps to 5 in our experiments.

    \item \textbf{Multi-Agent Debate} (MAD; \citet{liang-etal-2024-encouraging}) introduces multiple agents that exchange arguments, together with a judge that manages the debate process and derives the final answer. In our evaluation, we set the maximum number of debate rounds to 3 and set the temperature to 0.0 during the debate process.

    \item \textbf{SocialGaze} \cite{vijjini-etal-2024-socialgaze} is a multi-step prompting framework for improving the integration of human social norms in language models. The key idea is to verbalize the social situation from multiple perspectives before making the final judgment. 

    \item \textbf{SocialAgent} \cite{yuan-etal-2024-measuring} is a multi-agent framework introduced in the study of social norm understanding in LLMs. It uses interactions among agents to improve social norm reasoning. 

    \item \textbf{CulturalDebate} \cite{ki-etal-2025-multiple} extends multi-agent debate to culturally situated social norm reasoning. It utilizes two language model agents to discuss a cultural scenario and collaboratively reach a final decision. 

    \item \textbf{MetaMind} \cite{zhang2026metamind} is a meta-cognitive framework for modeling human social thoughts. It utilizes specialized agents to infer mental states, refine social hypotheses, and generate contextually appropriate responses. In our experiments, we set the maximum number of hypotheses to 7, the maximum number of revisions to 3, the maximum number of memory items to 100, and the temperature to 0.7.

\end{itemize}
We also include several representative training-based methods for a broader comparison.
\begin{itemize}[leftmargin=*]

    \item \textbf{SDPO} \cite{kong-etal-2025-sdpo} is a training-based method that selects key segments within multi-turn interactions for preference optimization. It aims to improve social agent behavior while reducing training noise. We train Qwen2.5-7B \cite{qwen2.5} based on the SDPO datasets\footnote{\url{https://huggingface.co/datasets/Tongyi-ConvAI/SDPO}} following the official training arguments.
    
    \item \textbf{SOTOPIA-$\Omega$} \cite{zhang-etal-2025-sotopia} transfers human social strategies into language agents through dynamic strategy injection. It uses negotiation-inspired reasoning strategies and direct strategies to construct high-quality social dialogue training data.

    \item \textbf{SOTOPIA-RL} \cite{yu2025sotopiarl} is a reinforcement learning framework for improving social intelligence in open-ended social interactions. It refines coarse episode-level feedback into utterance-level and multi-dimensional rewards, which helps improve credit assignment and reduce reward hacking.
    
\end{itemize}

\subsection{Details of Benchmarks}
\label{sec:appendix_details_of_benchmarks}
\boldpara{SOTOPIA Benchmark.} 
SOTOPIA evaluates artificial agents in goal-oriented social interactions through a multi-dimensional evaluation framework for social intelligence.
Each episode is scored after the interaction along seven dimensions as follows.

\begin{itemize}[leftmargin=*]

    \item \textbf{Believability \textsc{(Bel)} [0-10]} measures whether the agent behaves in a natural and realistic manner and whether its actions are aligned with its assigned character profile.

    \item \textbf{Relationship \textsc{(Rel)} [-5-5]} captures how the interaction affects the relationship between agents, including interpersonal relations, social status, and reputation.

    \item \textbf{Knowledge \textsc{(Kno)} [0-10]} evaluates the agent's ability to actively acquire new and useful information. 

    \item \textbf{Secret \textsc{(Sec)} [-10-0]} assesses whether the agent keeps its secret information or private intentions. A lower score indicates more severe leakage of private information.
    
    \item \textbf{Social Rules \textsc{(Soc)} [-10-0]} measures whether the agent follows social norms, legal rules, and other socially expected constraints.

    \item \textbf{Financial and Material Benefits \textsc{(Fin)} [-5-5]} evaluates both the agent's short-term monetary benefits and long-term economic payoffs.
    
    \item \textbf{Goal Completion \textsc{(Goal)} [0-10]} focuses on the extent to which an agent achieves its assigned social goal.

\end{itemize}

\boldpara{CEDAR Benchmark.}
CEDAR \cite{dai-etal-2026-tears} evaluates culturally grounded emotion alignment in multilingual and multimodal settings.
It covers seven languages and 14 fine-grained emotion categories.
It includes both multimodal and text-only subsets, with each language consisting of 400 multimodal data and 1,166 text-only data.
A multimodal instance consists of an image, a narrative, and a question in the target language, while the text-only one consists of a narrative and a question.
While inputs across languages are semantically equivalent, the ground-truth label is culture-specific and reflects the sociocultural norms associated with the target language.

\subsection{Efficiency of Our Framework}
\label{sec:appendix_efficiency_of_framework}
We assess the computational efficiency of our framework on the SOTOPIA benchmark, focusing on latency and token consumption. 
As illustrated in Figure~\ref{fig:latency_and_token}, our approach achieves lower average latency than the MAD and MetaMind baselines, facilitating more responsive interactions. 
Additionally, it consumes fewer tokens compared to MetaMind, demonstrating more economical resource usage.
This computational advantage primarily stems from concurrently processing the cultural inference and memory updates, thereby enhancing overall efficiency.

\subsection{Analysis of Dynamic Cultural Shifts.}
To further examine whether DyCAC particularly benefits interactions that require turn-level cultural adaptation, we stratify SOTOPIA instances according to the frequency of changes in the dominant cultural reference anchor.
We define an instance as \textbf{high-shift} if the dominant reference anchor changes more than twice across dialogue turns. 
Under this criterion, 70.4\% of the evaluated instances are categorized as high-shift.
As shown in Table~\ref{tab:shift_analysis}, the improvement brought by DyCAC is larger on the high-shift subset than on the low/no-shift subset, providing additional evidence that dynamic cultural calibration contributes more strongly in interactions with frequent profile revisions.

\begin{table}[t]
\centering
\small
\resizebox{\linewidth}{!}{
\begin{tabular}{lccc}
\toprule
\textbf{Subset} &
\textbf{Qwen2.5-7B} &
\textbf{Qwen2.5-7B + DyCAC} &
\textbf{$\Delta$} \\
\midrule
High-shift   & 0.7322 & 0.8182 & +0.0860 \\
Low/no-shift & 0.7043 & 0.7529 & +0.0486 \\
\bottomrule
\end{tabular}
}
\caption{Performance stratified by the frequency of dynamic cultural shifts on SOTOPIA.}
\label{tab:shift_analysis}
\end{table}

\section{Detailed Prompts}
\label{sec:appendix_details_of_prompts}

\begin{figure*}[t]
    \centering
    \begin{tcolorbox}[
        title={Prompts for Perception - Objective Fact Extraction},
        colback=gray!5,
        colframe=gray!75!black,
        fonttitle=\bfseries\color{white},
        coltitle=white,
        colbacktitle=gray!75!black,
        boxrule=0.8pt,
        arc=3mm,
        width=\textwidth 
    ]

    You are a dialogue analyst. 
        
    Extract objective facts about the latest counterpart move, using recent dialogue only as supporting context.
    
    Output this exact JSON shape (no extra keys):

    \begin{verbatim}
```JSON
{
  "static_attributes": {...},
  "dynamic_events": [...]
}
    \end{verbatim}
    static\_attributes: key-value pairs for stable user attributes (name, age, gender, occupation, location, relationships, explicit preferences). 
    Only include attributes that are explicitly mentioned. Omit null/unknown fields entirely.

    dynamic\_events: list of objects like \{"event": "...", "time\_reference": "..."\} for each event with a time anchor (e.g. "yesterday", "last week", "unspecified").

    \texttt{\{memory\_hint\}}

    Dialogue:
    \texttt{\{dialogue\}}

\end{tcolorbox}
\end{figure*}

\begin{figure*}[t]
    \centering
    \begin{tcolorbox}[
        title={Prompts for Perception - Mental State Analysis},
        colback=gray!5,
        colframe=gray!75!black,
        fonttitle=\bfseries\color{white},
        coltitle=white,
        colbacktitle=gray!75!black,
        boxrule=0.8pt,
        arc=3mm,
        width=\textwidth 
    ]

    You are a cognitive analyst. 
    
    Identify the speaker's mental state from dialogue. 
    Output ONLY valid JSON. 
    No preamble, no explanation, no markdown fences.

    Analyze the latest counterpart speaker's mental state. 
    Use recent dialogue only as context and focus on the latest-message section.

    Output this exact JSON shape (no extra keys):

    \begin{verbatim}
```JSON
{
  "immediate_intent": "<the speaker's direct goal or purpose in this turn>",
  "emotion": {
    "category": "<one word label, e.g. anxious, happy, frustrated, neutral>",
    "intensity": "<low | medium | high>"
  },
  "values_and_obsessions": ["<implicit deep belief or recurring concern>"]
}
    \end{verbatim}
    values\_and\_obsessions: list up to 3 implicit values inferred from the text. Return an empty list if none are detectable.

    Dialogue:
    \texttt{\{dialogue\}}

\end{tcolorbox}
\end{figure*}

\begin{figure*}[t]
    \centering
    \begin{tcolorbox}[
        title={Prompts for Perception - Cultural Cue Identification},
        colback=gray!5,
        colframe=gray!75!black,
        fonttitle=\bfseries\color{white},
        coltitle=white,
        colbacktitle=gray!75!black,
        boxrule=0.8pt,
        arc=3mm,
        width=\textwidth 
    ]

    You are a cross-cultural communication expert. 
    
    Identify cultural signals in dialogue. 
    Output ONLY valid JSON. 
    No preamble, no explanation, no markdown fences.
    
    Identify cultural communication patterns expressed by the latest counterpart speaker. Use prior turns only as context.

    Output this exact JSON shape (no extra keys):
    \begin{verbatim}
```JSON
{
  "communication_style": "<e.g. direct, indirect, formal, informal, assertive, 
  hedging>",
  "social_tendencies": ["<observable cultural pattern, e.g. collectivism, 
  hierarchy awareness>"]
}
    \end{verbatim}

    social\_tendencies: list up to 4 cultural behavioral patterns. Return an empty list if none are observable.

    Dialogue:
    \texttt{\{dialogue\}}

\end{tcolorbox}
\end{figure*}

\begin{figure*}[t]
    \centering
    \begin{tcolorbox}[
        title={Prompts for Cultural Profile Inference - Turn 1},
        colback=gray!5,
        colframe=gray!75!black,
        fonttitle=\bfseries\color{white},
        coltitle=white,
        colbacktitle=gray!75!black,
        boxrule=0.8pt,
        arc=3mm,
        width=\textwidth 
    ]

    You are a cultural analyst expert in Hofstede's dimensions.

    Based on the structured perception data extracted from the user's dialogue below, identify the {n} most likely countries or regions of cultural origin.
    
    Use communication style, social tendencies, values, emotional cues, behavioral patterns, and negotiation style as evidence.
    Prefer broad regions (e.g. Arab countries, Africa West) when the signal is coarse.
    Return exactly \texttt{\{n\}} country or region names ordered from most to least plausible.

    Perception data:
    \texttt{\{perception\_data\}}

    Output JSON only. No preamble, no markdown fences:
    \begin{verbatim}
```JSON
{"countries": ["<country>", ...]}
    \end{verbatim}
\end{tcolorbox}
\end{figure*}

\begin{figure*}[t]
    \centering
    \begin{tcolorbox}[
        title={Prompts for Cultural Profile Inference - Cultural Prior Scoring},
        colback=gray!5,
        colframe=gray!75!black,
        fonttitle=\bfseries\color{white},
        coltitle=white,
        colbacktitle=gray!75!black,
        boxrule=0.8pt,
        arc=3mm,
        width=\textwidth 
    ]

    You are a cross-cultural psychologist.
    
    Rate how plausible it is that the speaker originates from \texttt{\{country\}}, given the structured perception data below.
    Score from 0.0 (very implausible) to 1.0 (very plausible). Evaluate only this country independently.
    
    Perception data:
    \texttt{\{perception\_data\}}
    
    Output JSON only. No preamble, no markdown fences:
    \begin{verbatim}
```JSON
{"country": "{country}", "prior_score": <float 0.0-1.0>}
    \end{verbatim}
\end{tcolorbox}
\end{figure*}

\begin{figure*}[t]
    \centering
    \begin{tcolorbox}[
        title={Prompts for Cultural Profile Inference - Compatibility Score Calculation},
        colback=gray!5,
        colframe=gray!75!black,
        fonttitle=\bfseries\color{white},
        coltitle=white,
        colbacktitle=gray!75!black,
        boxrule=0.8pt,
        arc=3mm,
        width=\textwidth 
    ]

    You are a cross-cultural psychologist.
    
    Assume the speaker is from \texttt{\{country\}} (PDI=\texttt{\{PDI\}}, IDV=\texttt{\{IDV\}}, MAS=\texttt{\{MAS\}}, UAI=\texttt{\{UAI\}}, LTO=\texttt{\{LTO\}}, IVR=\texttt{\{IVR\}}).
    
    Dialogue:
    \texttt{\{dialogue\}}
    
    How naturally does this dialogue fit a speaker from \texttt{\{country\}}?
    Score 0.0 (very unnatural for this culture) to 1.0 (very natural). Evaluate this country independently.
    
    Output JSON only. No preamble, no markdown fences:
    \begin{verbatim}
```JSON
{"country": "{country}", "likelihood": <float 0.0-1.0>}
    \end{verbatim}
\end{tcolorbox}
\end{figure*}

\begin{figure*}[t]
    \centering
    \begin{tcolorbox}[
        title={Prompts for Cultural Profile Inference - Direct Style Inference},
        colback=gray!5,
        colframe=gray!75!black,
        fonttitle=\bfseries\color{white},
        coltitle=white,
        colbacktitle=gray!75!black,
        boxrule=0.8pt,
        arc=3mm,
        width=\textwidth 
    ]

    You are a cross-cultural psychologist using Hofstede-style dimensions as soft latent variables, not hard labels.
    
    Infer the speaker's probable style on each dimension from 0 to 100, where 50 means unclear / mixed.
    Use the perception summary and the latest dialogue jointly.

    Perception data:
    \texttt{\{perception\_data\}}
    
    Latest dialogue:
    \texttt{\{dialogue\}}
    
    Output JSON only. No preamble, no markdown fences:
    \begin{verbatim}
```JSON
{
  "PDI": <0-100>,
  "IDV": <0-100>,
  "MAS": <0-100>,
  "UAI": <0-100>,
  "LTO": <0-100>,
  "IVR": <0-100>
}
    \end{verbatim}

\textbf{Rules:}

- Use 40-60 when evidence is weak.

- Base scores on observable interaction style, not demographics.

- Do not output explanations.

\end{tcolorbox}
\end{figure*}

\begin{figure*}[t]
    \centering
    \begin{tcolorbox}[
        title={Prompts for ToM-Driven Memory - ToM Inference},
        colback=gray!5,
        colframe=gray!75!black,
        fonttitle=\bfseries\color{white},
        coltitle=white,
        colbacktitle=gray!75!black,
        boxrule=0.8pt,
        arc=3mm,
        width=\textwidth 
    ]
    
    You are a Theory-of-Mind reasoner embedded in a dialogue agent. 
    
    Given a memory snapshot and the latest dialogue perception, infer the user's hidden mental states. 
    Output ONLY valid JSON. 
    No preamble, no explanation, no markdown fences.
    
    prior\_memory:
    \texttt{\{memory\_json\}}
    
    perception:
    \texttt{\{perception\_json\}}
    
    Infer hidden mental states. Output this exact JSON shape (no extra keys):
    \begin{verbatim}
```JSON
{
  "user_model_of_ai": "<what the user believes the AI currently knows about them; 
  use 'unknown' on the first turn>",
  "hidden_desires": [
    {
      "desire": "<latent need NOT directly stated>",
      "confidence": 0.0,
      "evidence": "<brief rationale citing perception signals>"
    }
  ],
  "belief_revision_event": {
    "detected": false,
    "slot": null,
    "old_value": null,
    "new_value": null,
    "rationale": ""
  },
  "dominant_topic": "<2-4 word label for the main topic>"
}
    \end{verbatim}
    \textbf{Rules:}
    
    - hidden\_desires: include 1-3 items with confidence >= 0.50; return empty list if none.
    
    - belief\_revision\_event: set detected = true only when the user clearly contradicts a prior intent in memory.
    
    - Do NOT fabricate. Use null/false when something cannot be reliably inferred.

\end{tcolorbox}
\end{figure*}

\begin{figure*}[t]
    \centering
    \begin{tcolorbox}[
        title={Prompts for ToM-Driven Memory - Memory Operations},
        colback=gray!5,
        colframe=gray!75!black,
        fonttitle=\bfseries\color{white},
        coltitle=white,
        colbacktitle=gray!75!black,
        boxrule=0.8pt,
        arc=3mm,
        width=\textwidth 
    ]

    You are a memory operations agent. 
    
    Given a memory snapshot and dialogue perception, generate the minimal set of update operations needed. 
    Output ONLY valid JSON. No preamble, no explanation, no markdown fences.

    prior\_memory:
    \texttt{\{memory\_json\}}
    
    perception:
    \texttt{\{perception\_json\}}
    
    Generate memory update operations. Output this exact JSON shape (no extra keys):
    \begin{verbatim}
```JSON
{
  "memory_operations": [
    {
      "action": "<ASSERT | REVISE | RETRACT | HOLD>",
      "layer": "<world_facts | mental_state>",
      "bdi_key": "<beliefs | desires | intentions | null>",
      "slot": "<snake_case_slot_name>",
      "value": "<new value; required for ASSERT and REVISE>",
      "confidence": 0.0,
      "likelihood": 0.0,
      "rationale": "<one sentence justification>"
    }
  ]
}
    \end{verbatim}

Action semantics:

- ASSERT  : Add a new slot not yet in prior\_memory.

- REVISE  : Update an existing slot. likelihood = P(evidence | new\_value\_correct). Bayes rule is applied externally.

- RETRACT : Remove a slot that is false, completed, or superseded.

- HOLD    : Explicitly acknowledge no change is warranted.

\textbf{Rules:}

- world\_facts layer: stable factual attributes (name, job, location, relationships). Set bdi\_key to null.

- mental\_state layer: BDI slots only. bdi\_key MUST be exactly one of: beliefs, desires, intentions.

- Do NOT emit operations for layer "dialogue\_meta" — it is managed automatically.

- Be conservative: prefer HOLD over noisy assertions.

- Confidence values must reflect genuine epistemic uncertainty; avoid extremes.

\textbf{[CRITICAL]}

bdi\_key mapping (perception field names are NOT valid bdi\_key values):

  perception "values\_and\_obsessions"  ->  bdi\_key: "beliefs"   (user's deep convictions)
  
  perception "immediate\_intent"       ->  bdi\_key: "intentions" (what the user plans to do)
  
  perception "emotion"                ->  bdi\_key: "beliefs"    (user's belief about their situation)
  
  perception "communication\_style"    ->  layer: world\_facts,   bdi\_key: null
  
  perception "social\_tendencies"      ->  layer: world\_facts,   bdi\_key: null
  
  Never use "values\_and\_obsessions", "emotion", or "immediate\_intent" as the bdi\_key field.

\end{tcolorbox}
\end{figure*}

\begin{figure*}[t]
    \centering
    \begin{tcolorbox}[
        title={Prompts for Planning and Execution - Adaptive Response},
        colback=gray!5,
        colframe=gray!75!black,
        fonttitle=\bfseries\color{white},
        coltitle=white,
        colbacktitle=gray!75!black,
        boxrule=0.8pt,
        arc=3mm,
        width=\textwidth 
    ]
You are a culturally intelligent conversational assistant.

Relevant profile:
\texttt{\{memory\_summary\}}

Style calibration:
\texttt{\{cultural\_guidelines\}}

Respond to the latest message naturally and helpfully. Keep the reply concise, specific, and grounded in the conversation.
\end{tcolorbox}
\end{figure*}

\begin{figure*}[t]
    \centering
    \begin{tcolorbox}[
        title={Prompts for Planning and Execution - Strategic Planner},
        colback=gray!5,
        colframe=gray!75!black,
        fonttitle=\bfseries\color{white},
        coltitle=white,
        colbacktitle=gray!75!black,
        boxrule=0.8pt,
        arc=3mm,
        width=\textwidth 
    ]    
    You are the hidden strategist for a SOTOPIA dialogue agent.
    You never produce the final utterance. You produce a compact tactical plan for ONE turn.
    
    Role / character:
    \texttt{\{agent\_persona\}}
        
    Private goal:
    \texttt{\{social\_goal\}}
    
    Character-scene brief:
    \texttt{\{base\_context\}}
    
    Strategic state:
    \texttt{\{strategy\_summary\}}
    
    Turn policy prior:
    \texttt{\{turn\_policy\}}
    
    Cultural style calibration:
    \texttt{\{cultural\_guidelines\}}
    
    Latest interaction block:
    \texttt{\{current\_input\}}

    Output JSON only with this exact schema:
    \begin{verbatim}
```json
{
  "primary_objective": "<single-turn objective>",
   "tactic": <probe|align|reassure|trade|concede|reframe|close|boundary|exit>",
  "target_effect": "<what change in the partner you want this turn>",
  "partner_constraint": "<best guess of the main blocker or concern>",
  "relationship_guard": "<how to avoid relationship damage>",
  "secret_guard": "<how to avoid unnecessary secret leakage>",
  "social_rule_guard": "<how to stay norm-compliant>",
  "believability_anchor": "<detail that keeps the move in character and scenario>",
  "recommended_content": ["<point 1>", "<point 2>"],
  "avoid": ["<thing to avoid 1>", "<thing to avoid 2>"],
  "should_leave": false
}
    \end{verbatim}

\textbf{Rules:}

- Optimize for the SOTOPIA dimensions jointly: goal, believability, relationship, knowledge gain, secret protection, social rules, and material benefit when relevant.

- Prefer moves that create commitment, information gain, leverage, a narrowed ask, or a concrete next step.

- Do NOT recommend generic empathy-only replies.

- Keep recommended\_content and avoid short and concrete.

- Set should\_leave = true only if continuing is clearly harmful, futile, or socially impossible.

\end{tcolorbox}
\end{figure*}

\begin{figure*}[t]
    \centering
    \begin{tcolorbox}[
        title={Prompts for Main Pipeline Context Construction},
        colback=gray!5,
        colframe=gray!75!black,
        fonttitle=\bfseries\color{white},
        coltitle=white,
        colbacktitle=gray!75!black,
        boxrule=0.8pt,
        arc=3mm,
        width=\textwidth 
    ]
    
    \textbf{[Conversation context]}
    
    Recent dialogue:
    \texttt{\{history\_block\}}
    
    \textbf{[Latest message]}

    \texttt{\{current\_input\}}
    
    \textbf{Task:} Use prior turns only as context. Analyze and respond to the latest message only.    
    
\end{tcolorbox}
\end{figure*}

\end{document}